\documentclass{article}
\usepackage{ijcai22}

\usepackage{times}
\usepackage{soul}
\usepackage{url}
\usepackage[hidelinks]{hyperref}
\usepackage[utf8]{inputenc}
\usepackage[small]{caption}
\usepackage{graphicx}
\usepackage{amsmath}
\usepackage{amsthm}
\usepackage{booktabs}
\usepackage{amssymb}
\usepackage{algorithm}
\usepackage{algorithmic}
\title{On Tracking Dialogue State by Inheriting Slot Values in Mentioned Slot Pools}

\author{
Zhoujian Sun$^1$
\and
Zhengxing Huang$^2$\And
Nai Ding$^{2,1,}$\footnote{Corresponding author}
\affiliations
$^1$Zhejiang Lab\\
$^2$Zhejiang University
\emails
sunzj@zhejianglab.edu.cn, 
\{zhengxinghuang, nai\_ding\}@zju.edu.cn
}

\begin{document}

\maketitle

\begin{abstract}
    Dialogue state tracking (DST) is a component of the task-oriented dialogue system. 
    It is responsible for extracting and managing slot values according to 
    utterances, where each slot represents a part of the information to accomplish 
    a task, and slot value is updated recurrently in each dialogue turn. However, 
    many DST models cannot update slot values appropriately. These models may repeatedly 
    inherit wrong slot values extracted in previous turns, resulting in the fail of the 
    entire DST task.
    They cannot update indirectly mentioned slots well, either. This study designed 
    a model with a \textbf{\textit{mentioned slot pool}} (MSP) 
    to tackle the update problem. The MSP is a slot-specific memory that records 
    all mentioned slot values that may be inherited, and our model updates slot 
    values according to the MSP and the dialogue context. Our model rejects 
    inheriting the previous slot value when it predicates the value is wrong. 
    Then, it re-extracts the slot value from the current dialogue context. As the 
    contextual information accumulates, the new value 
    is more likely to be correct. It also can track the indirectly mentioned slot 
    by picking a value from the MSP. Experimental results showed our model reached 
    state-of-the-art DST performance on MultiWOZ datasets\footnote{We released the source 
    code of this paper at \url{https://github.com/ZJLAB-AMMI/msp}}.
\end{abstract}

\section{Introduction}

The task-oriented dialogue system is a type of system that aims to collect information according to a 
multi-turn 
dialogue between a user and an agent to accomplish a task. Dialogue state tracking (DST) is a module 
of the system 
that is responsible for extracting values from utterances to fill slots and maintaining slots over the 
continuation of the dialogue, where each slot represents an essential part of the information and  
turn-specific values of all slots comprise the dialogue state \cite{heck-2020-trippy,ni-2021-DSTReview}.

\begin{figure}[tb]
    \centering
    \includegraphics[width=8.1144cm,height=9.3024cm]{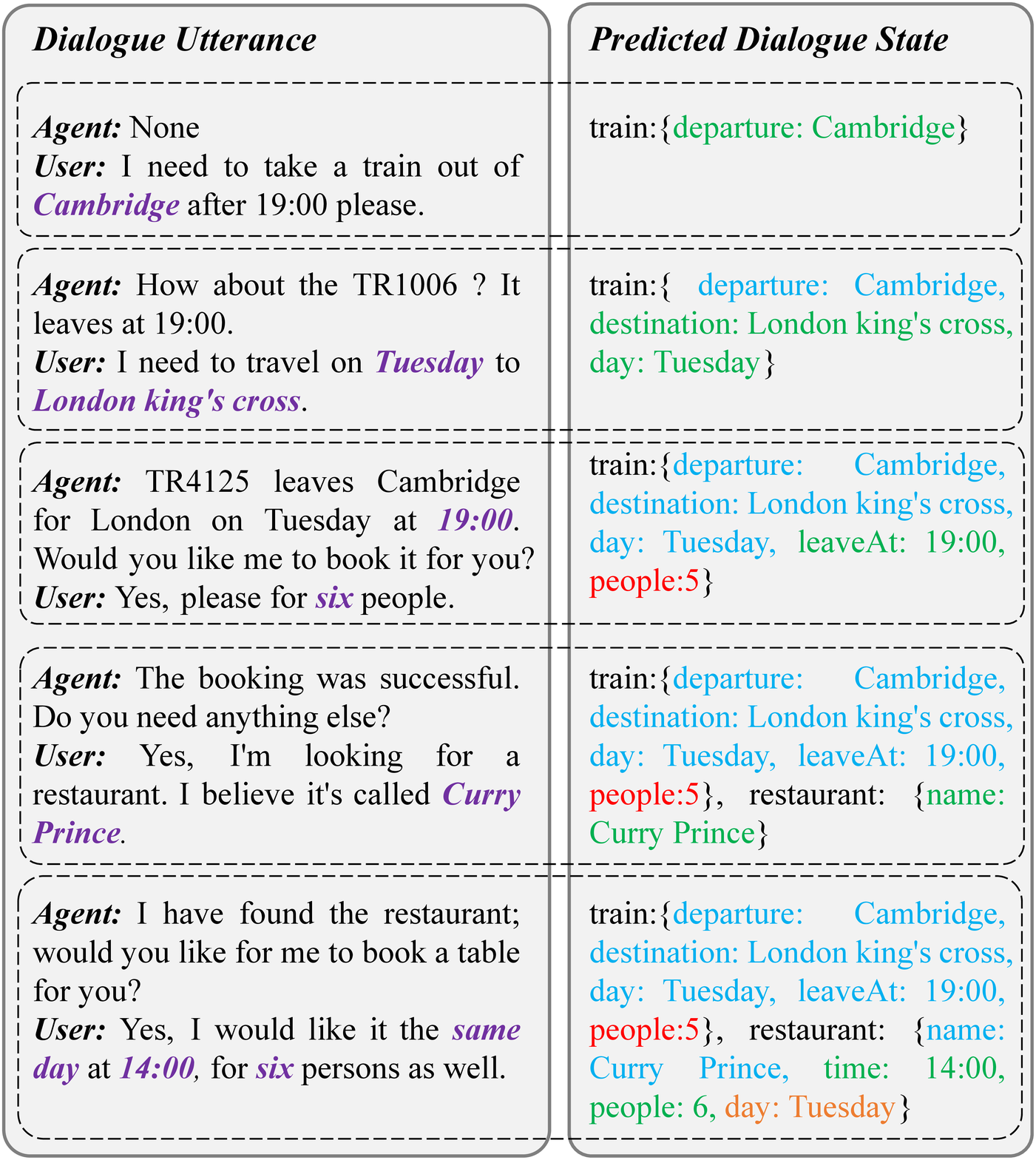}
    \caption{Sample DST process. Green, blue, red, and orange slots indicate 
    the value is updated via current turn utterances, inherited from the previous 
    turn, wrong, and from another slot, respectively. We used purple to mark key 
    information in utterances.}
    \label{fig:SampleDST}
\end{figure}

Figure \ref{fig:SampleDST} describes a sample DST process. As each slot is typically mentioned only 
once in the entire 
dialogue, the dialogue state is updated recurrently. Therefore, the dialogue state update strategy 
plays a critical role in the DST task. 
However, we found this topic is not detailly investigated. Many previous studies adopted a naïve update 
strategy that directly inherits the previous value when a slot is not mentioned in the current turn 
\cite{chao-2019-BERTDST}. Once a model extracts a wrong slot value, the wrong value may be repeatedly inherited 
in the following dialogue, resulting in the fail of the entire DST task, e.g., the train-people slot in the 
sample \cite{manotumruksa-2021-turnBaseLoss,zhao-2021-effectiveSequence}. 
Furthermore, a slot may be mentioned indirectly in a complex DST task as the value is referred from 
another slot rather than explicitly mentioned in current turn utterances 
\cite{zhou-2019-DSTQA,heck-2020-trippy}, e.g., the value of restaurant-day slot in the sample is from the 
train-day slot. An intelligent model needs to reject inheriting wrong values from previous 
turns and correctly track values for indirectly mentioned slots. Psychological studies have 
shown that humans can constantly monitor and update wrong interpretations during language processing. 
For example, when listening to the first a few words of a sentence, the listener will build a plausible 
interpretation. If this interpretation is inconsistent with later words, the brain will re-analyze the 
sentence and adopt a new interpretation that is consistent with all the input \cite{townsend-2001-sentence}. 
Here, we adopt a similar strategy that allows models to update slot values based on subsequent input.

This study designed a model with an additional \textbf{\textit{mentioned slot pool}} 
(MSP) module to 
tackle the dialogue state update problem more elaborately. MSP is a slot-specific memory including all 
slot values that are possible to be inherited. For each slot, our model will determine whether to 
inherit the previous value or extract the value from utterances according to dialogue context and the 
MSP. This design enables the model not to inherit the previous slot value when it predicates the value is 
wrong. Then, the model re-extracts the slot value from current dialogue context. As contextual 
information accumulates with dialogue progresses, the new value extraction process is more likely 
to find the right value and correct previous mistakes. For example, the last turn of the sample DST 
contains the utterance 
"six persons \textbf{\textit{as well}}." This contextual information helps the model realize that 
the values of train-people 
and restaurant-people slots should be the same. As the value of the restaurant-people slot is six, the 
wrong train-people value may be corrected in the new value extraction process. Meanwhile, our model can 
track indirectly mentioned slot values by picking a value in MSP because all relevant slot values are 
integrated into it.

We investigated the performance of our model on three representative DST datasets. 
The result showed that our model achieved state-of-the-art (SOTA) performance among DST models 
which were not 
trained by external datasets. Further analysis also indicated that our design is more efficient 
than other dialogue state 
update methods. We used the abbreviation MSP to denote both the pool and our model in 
the following content.

\section{Related Work}
Recently, fine-tuning large pretrained neural network language model (PNNLM) gradually becomes the de facto 
standard paradigm to tackle DST tasks \cite{devlin-2019-BERT}. 
For example, Mehri et al. \shortcite{mehri-2020-DialoGLUE} 
fine-tuned BERT \cite{devlin-2019-BERT} to track dialogue state. This type of studies demonstrated that 
DST performance could be significantly improved by simply using larger PNNLM. The potential of the prompt 
technique also inspired researchers to fulfill the DST task by giving model slot 
descriptions \cite{zang-2020-MultiWOZ22,liu-2021-promptSurvey}. Some studies demonstrated the efficiency of 
conducting 
data augmentation. Song et al. \shortcite{song-2021-DSTDataAugment} and Summerville et al.
\shortcite{summerville-2020-DSTDataAugment} augmented data by copying utterances and 
replacing the slot value label. Li et al. \shortcite{li-2020-coco} used the pretrained utterance generator and 
counterfactual goal generator to create novel user utterances.

Meanwhile, another series of studies try to improve DST performance by designing a more effective model 
structure. Traditional DST models formulate slot filling as a classification task, requiring a predefined 
ontology containing all possible classification values \cite{nouri-2018-ScalableDST}. However, these models 
suffer from generalization issues. To solve this issue, Wu et al. \shortcite{wu-2019-TRADE} adopted an 
encoder-decoder 
framework to formulate the DST as a machine translation task, and Gao et al. \shortcite{gao-2019-DSTReadComprehension} 
formulated DST as a span finding task. Both methods are widely adopted in subsequent studies, e.g., 
\cite{tian-2021-AGDST,zhou-2019-DSTQA}. 
Previous studies also realized that the slot value might be 
mentioned indirectly. Heck et al. \shortcite{heck-2020-trippy}, Kim et al. \shortcite{kim-2020-OverwriteMemory}, 
and Zhou et al. \shortcite{zhou-2019-DSTQA} proposed a triple copy 
strategy, a selective overwrite method, and a knowledge evolving graph to deal with the indirect mention 
problem, respectively. Manotumruksa et al. \shortcite{manotumruksa-2021-turnBaseLoss} noticed the wrong slot 
value is mistakenly inherited, and they tackle this problem by amplifying the loss weight of DST on early turns. 
Although these studies have tried to solve the mistakenly inherit problem and the indirectly mention problem 
independently, none of them try to solve two problems at once, while we achieved this goal by introducing 
the model equipped with a MSP.

\begin{figure}[tb]
    \centering
    \includegraphics[width=8.144cm,height=6.93cm]{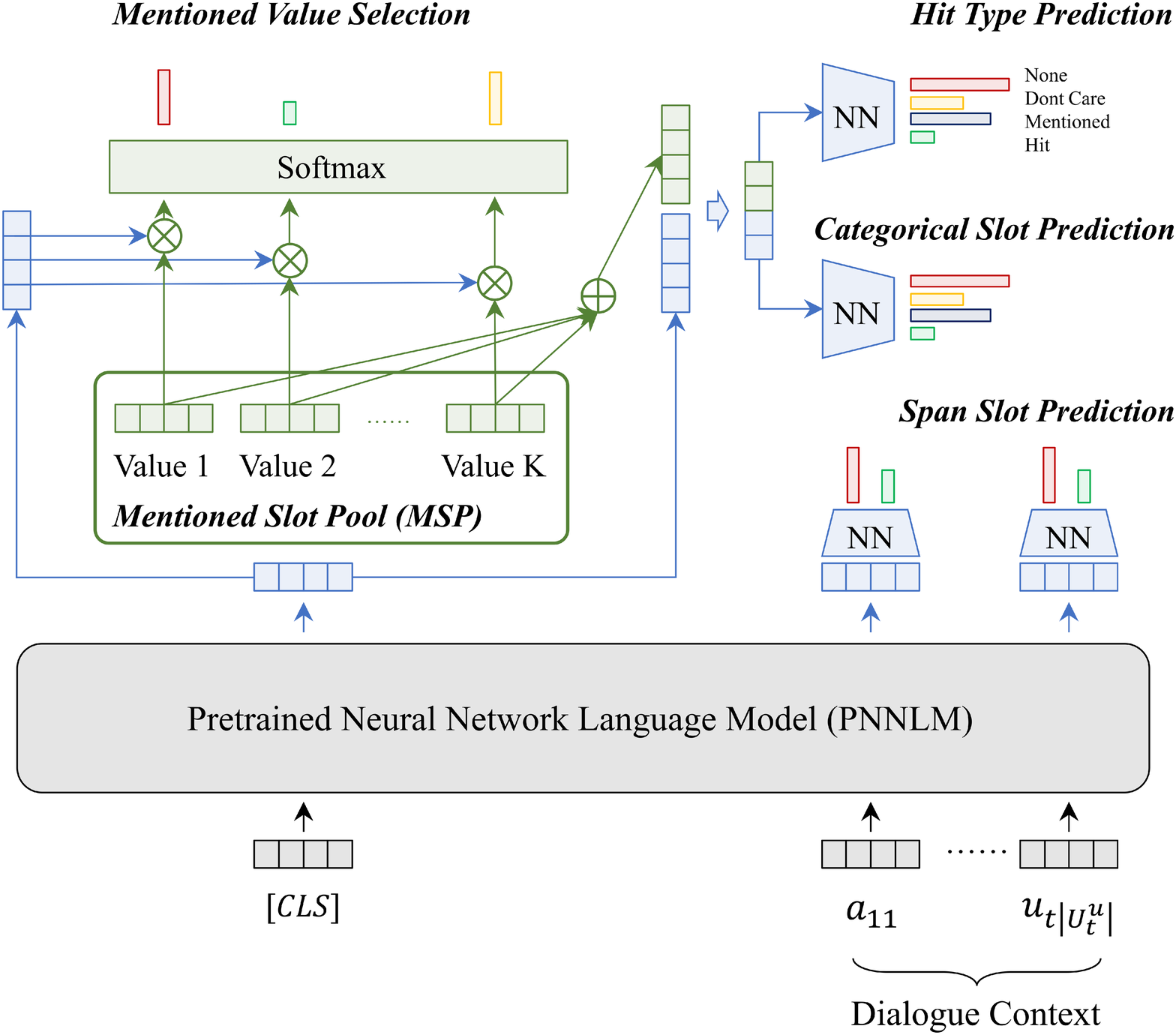}
    \caption{Model diagram}
    \label{fig:modelDiagram}
\end{figure}

\section{Methodology}
\subsection{Preliminaries}
Figure \ref{fig:modelDiagram} depicts the structure of our model. We represent a dialogue as 
$X=\{U_{1}^a,U_{1}^u,…,U_{T}^a,U_T^u \}$, where $T$ is the total turn number, 
$U_t^a$ and $U_t^u$ are utterances of agent and user in turn $t$, respectively. 
$U_t^a$ and $U_t^u$ consist of two lists of word tokens $a_{ti}$ and $u_{tj}$, 
respectively. 
Dialogue context $C_t=\{U_{1}^a,U_{1}^u,…,U_t^a,U_t^u\}$ indicates observable 
dialogue utterances at turn $t$. Following the setting of previous work \cite{heck-2020-trippy}, we add
a classification token $[CLS]$ in front of $C_t$, and feed this extended dialogue context into a PNNLM.
We use $R_t=[r_t^{CLS},r_t^{1},…,r_{t}^{\left\lvert C_t\right\rvert }]$ 
to denote the output of PNNLM, 
where $r_t^i \in \mathbb{R}^{n}$ corresponds to a token in $C_t$ and $n$ represents the output dimension of the PNNLM. 
The goal of DST is to exactly predict the 
dialogue state $y_t$ according to $C_t$. $y_t$ is a set of slot-value tuples. 

We use $\mathcal{M}_{s,t}=[M_{s,t}^{1},...,M_{s,t}^K]$ to denote the MSP, where $K$ is the maximum size of MSP. 
$M_{s,t}^i$ indicates the predicted values of slot $s$ or relevant slots $s^\prime$ at turn $t-1$. 
The definition of relevant slots is described later.
$m_{s,t}\in \mathbb{R}^{K\times n}$ and $m_{s,t}^i$ indicate the representation of $M_{s,t}$ and $M_{s,t}^i$, 
respectively. The low-rank bilinear model is utilized to generate a 
fused MSP representation \cite{kim-2018-bilinear},
\begin{equation}
    m_{s,t}^{\rm fused} = {\rm softmax}([r_{slot}+r_t^{CLS}]W_s^{\rm fused} m_{s,t}^T)m_{s,t}
    \label{eq:fuse}
\end{equation}

\noindent where $r_{slot}\in \mathbb{R}^{n}$ are representations of a given slot, and $W_s^{\rm fused}\in \mathbb{R}^{n\times n}$ 
is a learnable parameter.

\subsection{Hit Type Prediction}
Each slot is equipped with a hit type prediction layer. At each turn $t$, the hit type prediction layer maps 
representations of MSP and dialogue context
to one of the four classes in $\{none,dontcare,mentioned,hit\}$,

\begin{equation}
    p_{s,t}^{\rm type} = {\rm softmax}(W_s^{\rm type}[m_{s,t}^{\rm fused}+r_t^{CLS}]+b_s^{\rm type})\in \mathbb{R}^4
    \label{eq:hitType}
\end{equation}

\noindent where \textit{none} indicates the slot is not mentioned until turn $t$, \textit{dontcare} indicates the user does not care 
about the value of 
slot $s$, \textit{mentioned} indicates slot value is from an item in MSP, and 
\textit{hit} indicates slot value needs to be updated according 
to $C_t$. If a slot is already mentioned and the predicted slot hit type is \textit{hit}, it indicates our model 
predicts the previous 
slot value is wrong, and the model will update the slot value via the hit value prediction module.

\subsection{Mentioned Value Selection}
As described in equation \ref{eq:mentionedValueSelect}, 
we utilized a bilinear model to select the most appropriate slot value in MSP according to the 
representation of $C_t$ when our model assigns \textit{mentioned} as hit type. 
The value with biggest $p_{s,t}^{\rm mention}$ will be selected.

\begin{equation}
    p_{s,t}^{\rm mention} = {\rm softmax}(r_t^{CLS}W_s^{\rm mention}m_{s,t}^T)
    \label{eq:mentionedValueSelect}
\end{equation}

\subsection{Hit Value Prediction}
Our model extracts a slot value from $C_t$ when the model assigns \textit{hit} as hit type.
In this study, we refer to slots whose possible value number are small as categorical slots, 
e.g., whether a hotel has free internet, and slots whose possible value numbers are large, unenumerable, or may 
change over time as span slots, e.g., restaurant name in a booking task. The value of a categorical slot is predicted via 
a classification method. A slot-specific prediction layer takes $r_t^{CLS}$ and $m_{s,t}^{\rm fused}$
as input and generate the probabilities 
of each slot value,

\begin{equation}
    p_{s,t}^{\rm hit} = {\rm softmax}(W_s^{\rm hit}[m_{s,t}^{\rm fused}+r_t^{CLS}]+b_s^{\rm hit})\in 
    \mathbb{R}^{\left\lvert V_s\right\rvert}
    \label{eq:categoricalHitValue}
\end{equation}

\noindent where $V_s$ denotes the ontology of a categorical slot. We predict the value of a span slot by 
finding a token span within $C_t$. 
Our model determines the token span by predicting its start token index and end token index. 
A slot-specific span prediction layer takes $R_t$ as input and projects it as:
\begin{equation}
    [\alpha_{s,t}^i,\beta_{s,t}^i]=W_s^{\rm hit}r_t^i + b_s^{\rm hit} \in \mathbb{R}^2
    \label{eq:spanProbability}
\end{equation}
\begin{equation}
    p_{s,t}^{\rm start} = {\rm softmax}(\alpha_{s,t})
    \label{eq:spanStartIndex}
\end{equation}
\begin{equation}
    p_{s,t}^{\rm end} = {\rm softmax}(\beta_{s,t})
    \label{eq:spanEndIndex}
\end{equation}
The index with the biggest probability will be assigned as the classify value, start index, or end index. 
The span will be assigned as none if the start index is larger than the end index.

\subsection{Optimization}
The loss function for the hit type prediction, mentioned value selection, and hit value prediction of a single 
dialogue are defined as 
follows:
\begin{equation}
    \mathcal{L}_{\rm type}=\sum_{t}^{T}\sum_s^{\mathcal{S}}{-{\rm log}(y_{s,t}^{\rm type}(p_{s,t}^{\rm type})^T)}
    \label{eq:typeLoss}
\end{equation}
\begin{equation}
    \mathcal{L}_{\rm mention}=\sum_{t}^{T}\sum_s^{\mathcal{S}}{-{\rm log}(y_{s,t}^{\rm mention}(p_{s,t}^{\rm mention})^T)}
    \label{eq:mentionLoss}
\end{equation}
\begin{equation}
    \mathcal{L}_{\rm hit}=\sum_{t}^{T}\sum_s^{\mathcal{S}}
    \left\{
        \begin{array}{lr}
            -{\rm log}(y_{s,t}^{\rm hit}(p_{s,t}^{\rm hit})^T) \ (\rm categorical \ slot) \\
            -\frac{1}{2}({\rm log}(y_{s,t}^{\rm start}(p_{s,t}^{\rm start})^T) + \\ 
            {\rm log}(y_{s,t}^{\rm end}(p_{s,t}^{\rm end})^T)) \ (\rm span \ slot) \\
        \end{array}
    \right.
    \label{eq:hitLoss}
\end{equation}

\noindent where $y_{s,t}^{\rm type}$, $y_{s,t}^{\rm mention}$, $y_{s,t}^{\rm hit}$, $y_{s,t}^{\rm start}$, 
$y_{s,t}^{\rm end}$ are one-hot encoded labels of a slot hit type, mentioned slot, categorical slot, and the start index and the end 
index of a span slot, respectively. 
The joint loss function of dialogue is a weighted sum of $\mathcal{L}_{\rm type}$, $\mathcal{L}_{\rm mention}$, 
and $\mathcal{L}_{\rm hit}$, as shown in equation \ref{eq:jointLoss}, where $\alpha$, $\beta$, $\gamma$ are weight hyperparameters.

\begin{equation}
    \mathcal{L}=\alpha\mathcal{L}_{\rm type}+\beta \mathcal{L}_{\rm mention}+\gamma \mathcal{L}_{\rm hit}
    \label{eq:jointLoss}
\end{equation}

\section{Experiments}
\subsection{Experiment Settings}
\subsubsection{Dataset}
We conducted experiments on three annotated DST datasets, i.e., MultiWOZ 2.1, MultiWOZ 2.2, 
and WOZ 2.0, respectively \cite{eric-2020-multiwoz21,zang-2020-MultiWOZ22,wen-2017-WOZ2}.
We preprocessed datasets following \cite{heck-2020-trippy}.
We mainly focus on analyzing the results of MultiWOZ 2.1 and 2.2 because they 
are by far the most challenging open-source datasets in DST task.
MultiWOZ 2.1 and 2.2 
are comprised of over 10,000 multi-domain dialogues over a large ontology. There are five 
domains (train, restaurant, hotel, taxi, attraction), with 30 domain-slot pairs appearing 
in all data portions. We also report experimental results on the WOZ 2.0 to add additional 
evidence, although the it is smaller than the MultiWOZ dataset in both ontology 
and the number of examples.

\subsubsection{Range of Mentioned Slot Pools}
For a slot $s$ at turn $t$, the MSP is comprised of the value of slot $s$ and values of 
(at most) other three relevant slots $s^{\prime}$ at turn $t-1$. The none slot value is not included.
We define the $s^{\prime}$ is 
a relevant slot of $s$ if $s$ may inherit the value of slot $s^{\prime}$. Of note, a slot  
only inherit the value from a small fraction of other slots. For example, the taxi-destination 
slot cannot inherit the value from the restaurant-food slot and taxi-departure slot because 
the restaurant-food is not a place, and the destination cannot be the same as the departure. 
We designed a custom dictionary in this study to define the range of relevant slots. 
The MSP will be padded if its actual size is less than four. We used 
the masking method to avoid the model selecting the padded value. The MSP will be truncated if 
its actual size is larger than four. Only the latest four updated slot values will be reserved. 
If the actual size of MSP is zero and our model assigns the slot hit type as mentioned, 
the slot value will be assigned as none.

\subsubsection{Evaluation Metric}
We mainly evaluated DST models using the Joint Goal Accuracy (JGA) metric. Turn-specific
JGA is one if and 
only if all slot-value pairs are correctly predicted, otherwise zero. The general 
JGA score is averaged across all turns in the test set. 

Although JGA is the most widely used metric in the DST task, it is not comprehensive enough 
because the label distribution in the DST dataset is highly imbalanced. We adopted precision, 
recall, and F1 to investigate model performance more detailly. As slot filling is not a binary 
classification task, we define $\rm precision=\frac{TP}{(TP+FP)}$, $\rm recall=\frac{TP}{(TP+FN+PLFP)}$
, and F1 is the 
harmonic mean of recall and precision. TP (true positive) indicates the number of cases 
that the slot value is not none, and the model successfully predicts the value. FP (false positive) 
indicates that the slot value is none, but the model predicts not none. 
FN (false negative) indicates that the slot value is not none, but the model 
predicts none. PLFP (positive label false prediction) indicates that the slot value is not none 
and the model predicts a wrong positive value.

\subsubsection{Implemention Details}
We used the pre-trained BERT transformer as the PNNLM backbone \cite{devlin-2019-BERT}, 
which was also adopted in most previous DST studies. The base version of BERT was trained 
on lower-uncased English text. It has 12 hidden layers with 768 units and 12 self-attention heads. 
The large version has 24 hidden layers with 1024 units and 16 self-attention heads, and it was 
trained on cased English text. The base and large versions of BERT have about 110 million and 345 
million parameters, respectively. Unless specified, we used the 
base version of BERT as the pre-trained backbone and reported corresponding performance.

The maximum input sequence length was set to 512 tokens after tokenization. The weights $\alpha$, $\beta$, 
and $\gamma$ were 0.6, 0.2, and 0.2, respectively. We adopted embeddings released from WordPiece 
as value representations and slot representations $(m_{s,t}^i,r_{slot})$ \cite{wu-2016-google}. 
The word embeddings were locked during the training process. If the slot and the value need to be 
represented by multi-tokens, we used the mean of the corresponding token embeddings as the representation.

For optimization, we used Adam optimizer \cite{kingma-2014-adam}. The initial learning rate was set to 
1e$-$5, and the total epoch number was set to 20. We conducted training with a warmup proportion of 
10\% and let the learning rate decay linearly after the warmup phase. Early stopping was employed 
based on the JGA of the development set. All the reported performance JGA were the mean of five 
independent experiments.

\subsubsection{Baseline Models}
We compared our proposed model with a variety of recent DST baselines. 

\begin{itemize}
\item TRADE \cite{wu-2019-TRADE} encodes the whole dialogue context using bidirectional Gated Recurrent Units (GRU) 
and generates the value for every slot using the GRU-based copy mechanism.
\item SUMBT \cite{lee-2019-sumbt} learns the relations between domain-slot-types and slot-values appearing in 
utterances through attention mechanisms based on contextual semantic vectors. 
\item DS-DST \cite{zhang-2020-dual} is an ontology-based DST model that requires an ontology with 
all possible values for each domain-slot pair. 
\item Trippy \cite{heck-2020-trippy} uses the triple copy mechanism to track the dialogue state. 
\item Seq2Seq-DU \cite{feng-2021-sequence} employs two encoders to encode the utterances and the descriptions 
of schemas and a decoder to generate pointers to represent the state of dialogue.
\item AG-DST \cite{tian-2021-AGDST} generates a dialogue based on the current turn and the previous 
dialogue state and a two-pass process.
\end{itemize}

As our model is fine-tuned on the target dataset, we did not include models trained by augmented or external 
corpus as baselines to make the comparison fairly, e.g. \cite{mehri-2020-DialoGLUE,li-2020-coco}. 
The performance of baselines was cited from corresponding papers or \cite{zhao-2021-effectiveSequence}.

\subsection{Experimental Results}
\subsubsection{DST Performance}
Table 1 describes the DST performance of our MSP models and baselines in MultiWOZ 2.1, MultiWOZ 2.2, 
and WOZ 2.0 datasets, respectively. The domain-specific JGAs of two MultiWOZ datasets are described in Table 2. 
The MSP-B indicated the model used base version of BERT as the backbone, while the MSP-L indicated 
the model used the large version of BERT. The AG-DST-S and AG-DST-T indicates the two models 
used single PNNLM and two PNNLMs as backbones, respectively. The doamin-specific JGA indicated our MSP model obtained
better performance in taxi, resaurant, and attraction task because of the update of MultiWOZ dataset.

\begin{table}[tbh]
    \centering
    \begin{tabular}{lcccc}
    \toprule
      &  \textbf{\# of} & \multicolumn{2}{c}{\textbf{MultiWOZ}} & \textbf{WOZ} \\ 
    \textbf{Model}  & \textbf{Para.} & \textbf{2.1} & \textbf{2.2} & \textbf{2.0} \\
    \midrule
    \textbf{TRADE}  & / & 45.6\% & 45.4\% & /  \\
    \textbf{SUMBT}  & 110M & 49.2\% & 49.7\% & 91.0\% \\
    \textbf{DS-DST}  & 110M & 51.2\% & 51.7\% & 91.2\% \\
    \textbf{Trippy}  & 110M & 55.3\% & 50.7\% & \textbf{92.7\%} \\
    \textbf{MSP-B}  & 110M & \textbf{56.2\%} & \textbf{54.2\%} & 91.2\% \\
    \midrule
    \textbf{Seq2Seq-DU}  & 220M & 56.1\% & 54.4\% & / \\
    \textbf{AG-DST-S}  & 340M & / & 56.2\% & / \\
    \textbf{AG-DST-A}  & 680M & / & 57.1\% & / \\
    \textbf{MSP-L}  & 345M & \textbf{57.2\%} & \textbf{57.7\%} & / \\
    \bottomrule
    \end{tabular}
    \caption{Main results}
    \label{table:mainResult}
\end{table}

\begin{table}[h]
    \centering
    \begin{tabular}{lcc}
    \toprule
    \textbf{Model} & \textbf{MultiWOZ 2.1} & \textbf{MultiWOZ 2.2} \\
    \midrule
    \textbf{Taxi} & 96.2\% & 97.5\%   \\
    \textbf{Restaurant}  & 88.4\% & 88.8\% \\
    \textbf{Hotel}  & 85.0\% & 82.2\% \\
    \textbf{Attraction}  & 89.1\% & 89.3\% \\
    \textbf{Train}  & 89.5\% & 87.6\% \\
    \bottomrule
    \end{tabular}
    \caption{Domain-specific JGA of MSP}
    \label{table:DomainJGA}
\end{table}

As the size of PNNLM significantly influences the performance of models in almost all natural language 
processing tasks, it is necessary to figure out whether the performance improvement of a model is from 
its structure design or its PNNLM scale.  Therefore, we also described the number of parameters in PNNLM. 
The result showed that our MSP-B model achieved better performance than baselines when their PNNLM sizes were 
similar. Specifically, the MSP-B model improved SOTA JGA of MultiWOZ 2.1 from 55.3\% to 56.2\% (compared to Trippy) 
and MultiWOZ 2.2 from 51.7\% to 54.2\% (compared to DS-DST). It also achieved comparable performance (JGA: 91.2\%) 
compared to DS-DST and SUMBT in WOZ 2.0, though slightly worse than Trippy (JGA: 92.7\%).

Our MSP model is also more efficient than baselines because it achieved comparable or better performance with 
significantly fewer parameters and without utilizing the slot description information. Specifically, the 
MSP-B model obtained 56.2\% and 54.2\% JGA in two MultiWOZ datasets via only about 110 million parameters 
(one uncased-base BERT). The Seq2Seq-DU achieved similar performance via about 220 million parameters 
(two uncased-base BERTs) and the schema descriptions (JGA: 56.1\% and 54.4\% in two MultiWOZ datasets). 
Similarly, the MSP-L model achieved significantly better performance than AG-DST (JGA: 57.7\% vs. 56.2\% 
in MultiWOZ 2.2) when using PNNLMs with a similar number of parameters. The AG-DST model is slightly 
worse than our MSP model. Even it uses two times more parameters (JGA: 57.7\% vs. 57.1\% in MultiWOZ 2.2 dataset). 
Meanwhile, our MSP-L model achieved 57.2\% JGA in MultiWOZ 2.1 dataset. As far as we know, our MSP model reached 
a new SOTA in the MultiWOZ dataset among models not trained by external or augmented datasets.

\begin{table}[h]
    \centering
    \begin{tabular}{lccc}
    \toprule
    \textbf{Model} & \textbf{MultiWOZ 2.1} & \textbf{MultiWOZ 2.2} \\
    \midrule
    \textbf{Pure context}  & 53.7\% & 52.3\%   \\
    \textbf{Changed state}   & 54.9\% & 53.2\% \\
    \textbf{Full state} & 55.5\% & 53.6\% \\
    \textbf{MSP}  & \textbf{56.2\%} & \textbf{54.2\%} \\
    \bottomrule
    \end{tabular}
    \caption{Update strategy comparison}
    \label{table:RecurrentStrategyComparison}
\end{table}

\subsubsection{Update Strategy Comparison}

We conducted experiments on our strategies and three common strategies to investigate whether our MSP-based 
dialogue update strategy is better. The three strategies are:

\begin{itemize}
    \item Pure context strategy. This strategy does not use the previous dialogue state and tracking the 
    dialogue state purely relies on dialogue context. It is widely used in end-to-end 
    models, e.g., \cite{hosseini-Asl-2020-SimpleTOD}.
    \item Changed state strategy.
    This strategy utilizes the entire dialogue context to track slots changed in the latest turn. If a 
    slot is not mentioned in the latest turn, it inherits the value recorded in the previous dialogue state. 
    Heck et al. \shortcite{heck-2020-trippy} and Zhang et al. \shortcite{zhang-2020-dual} used this strategy.
    \item Full state strategy. This strategy converts previous dialogue state into a string, and utilizes the 
    dialogue context and dialogue state string to track entire dialogue state. We adopted the design of 
    AG-DST to implement this strategy \cite{tian-2021-AGDST}.
\end{itemize}

Table \ref{table:RecurrentStrategyComparison} describes the result of the dialogue state update strategy comparison, 
where all other experimental settings are the same. It is not surprising that the performance of the changed state strategy 
is better than the pure context strategy (JGA: 54.9\% vs. 53.7\% in MultiWOZ 2.1 and 53.2\% 
vs. 52.3\% in MultiWOZ 2.2) as dialogue state is a compact representation of the dialogue history. Moreover, 
our strategy achieved about 2\% and 1\% improvement compared to changed state strategy and full state strategy
as it achieved JGA as 56.2\% 
and 54.2\% in MultiWOZ 2.1 and 2.2 datasets, respectively. These results demonstrated that our MSP-based dialogue 
state update strategy is more effective in DST tasks.

\subsubsection{Ablation Study}
We conducted ablation studies to investigate the performance of our model in five different structures.

\begin{itemize}
    \item Span-short context. All slots are predicted via the span-based method, and the model tracks the 
    dialogue state purely based on the latest 128 tokens in the dialogue context.
    \item Long context. The model tracks dialogue state based on the latest 512 tokens.
    \item Categorical slots. The categorical slot is predicted via the classification-based method in this structure.
    \item MSP-self. Adding the MSP module into the model. Only the previous value of the target slot is included 
    in the MSP.
    \item MSP-full. Our design. Previous values of the target slot and relevant slots are included in the MSP.
\end{itemize}

Table 4 describes the result of the ablation study. The model with a long context reached significantly better 
performance than the model with a short context, demonstrating that DST models benefit from longer 
contextual information. 
Applying the classification-based method to track categorical slots also improves DST performance. 
These findings accord with findings of previous studies \cite{tian-2021-AGDST,zhou-2019-DSTQA}. 

The DST model obtained about 2\%-3\% performance improvement by equipping the MSP module. The MSP-full model 
obtained faint performance improvement compared to the MSP-self model (JGA: 56.2\% vs. 56.0\%, 54.2\% vs. 53.9\% 
in two MultiWOZ datasets, respectively). On the one hand, these results showed the effectiveness of updating the 
dialogue state via our MSP-based strategy. On the other hand, it indicates that integrating the value of another 
slot into the MSP is helpful, though the performance gain is not significant. The ablation study demonstrated 
that the MSP module could be used as an additional model component to improve the DST performance.

\begin{table}[t]
    \centering
    \begin{tabular}{lccc}
    \toprule
    \textbf{Model} & \textbf{MultiWOZ 2.1} & \textbf{MultiWOZ 2.2} \\
    \midrule
    \textbf{Span-short context}  & 46.4\% & 47.9\% \\
    \textbf{\ +Long context}  & 52.8\% & 52.0\% \\
    \textbf{\ \ +Categorical slots} & 54.6\% & 53.3\% \\
    \textbf{\ \ \  +MSP-self} & 56.0\% & 53.9\% \\
    \textbf{\ \ \ \ +MSP-full}  & \textbf{56.2\%} & \textbf{54.2\%} \\
    \bottomrule
    \end{tabular}
    \caption{Abalation study}
    \label{table:AbalationStudy}
\end{table}

\subsubsection{Inherit Analysis}
\begin{figure}[t]
    \centering
    \includegraphics[width=8cm,height=4.74cm]{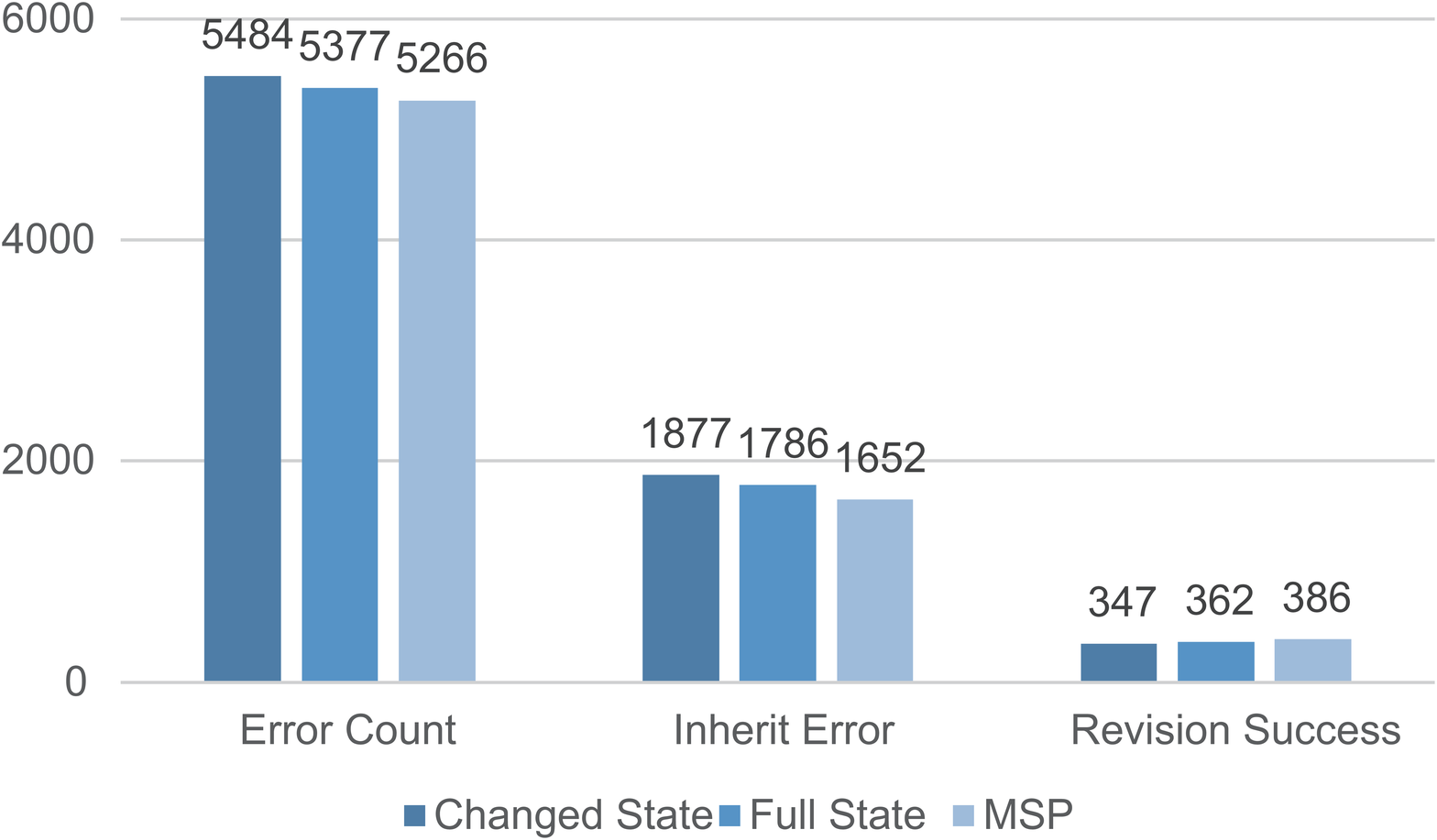}
    \caption{Inappropriate inherit analysis}
    \label{fig:errorComparison}
\end{figure}

\begin{figure}[t]
    \centering
    \includegraphics[width=8cm,height=4.74cm]{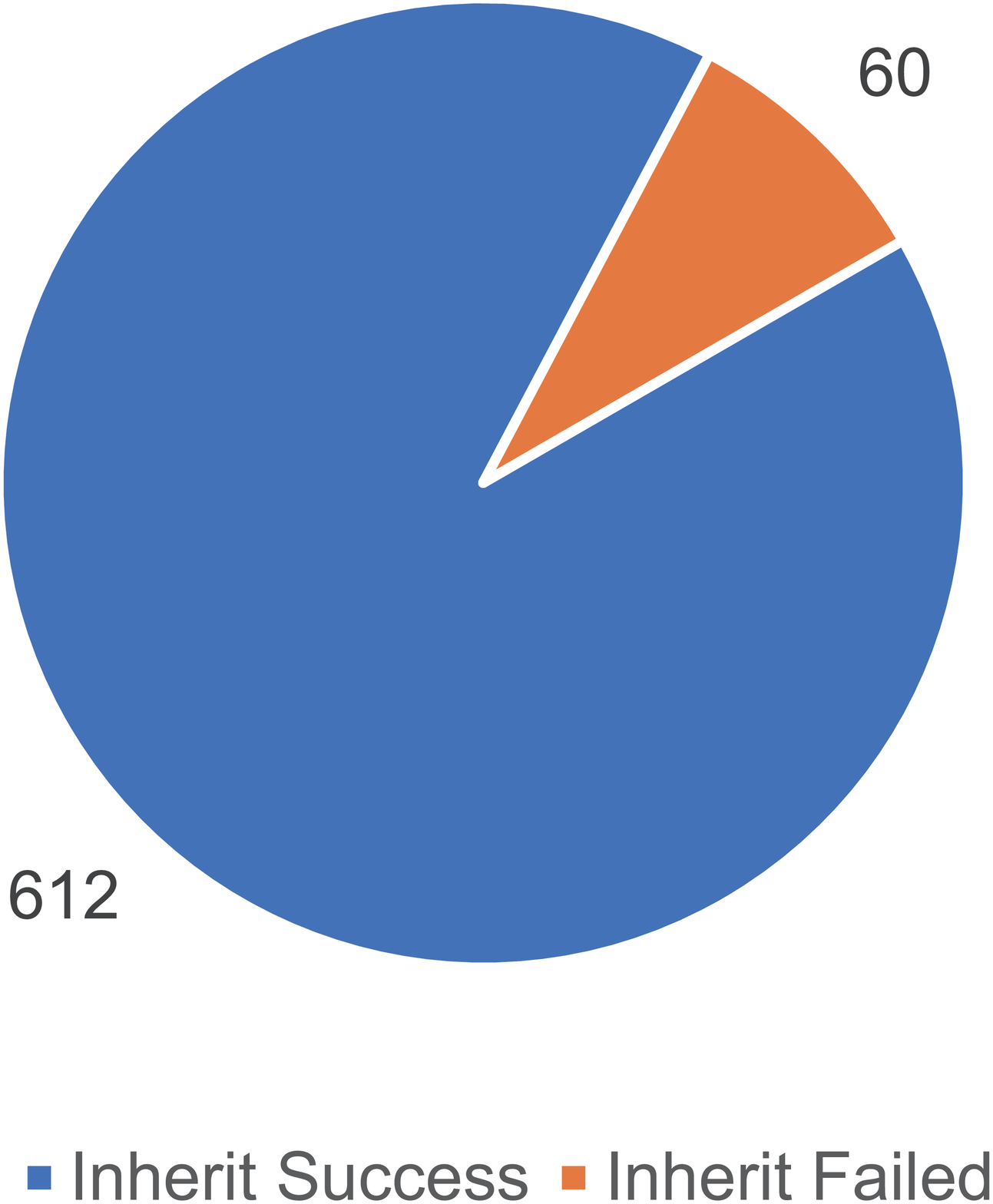}
    \caption{Indirect mentioned slot inherit analysis}
    \label{fig:indirectInherit}
\end{figure}

The previous three subsections have demonstrated the effectiveness of our model. 
In this subsection, we will further investigate the correctness of our assumption. 
That is, whether the MSP module improves DST performance by rejecting inheriting 
wrong slot values and tracking the indirectly mentioned slots. 

Figure \ref{fig:errorComparison} describes 
inappropriate inherit analysis result of the MSP, changed state, and full state based models on an experiment 
conducted in the MultiWOZ 2.2 dataset. Error count means the number of wrong slot value prediction 
cases. Inherit error means the error is caused by inappropriate inheriting. We defined inheriting the 
wrong previous slot value or failing to track indirectly mentioned slots as inappropriate inheriting. 
Revision success indicates the model rejects inheriting a wrong value and revising it into a correct 
value. The MSP model achieved better performance as it mistakenly predicted 5,266 slots, 218 times less 
than the changed state model. Meanwhile, the inappropriate inherit count of MSP is also less than 
the changed state model about 200 times. The MSP model successfully corrected 
the previous mistake 386 times, about 10\% more than the changed state model. 
These results indicate that performance improvement of the 
MSP model is likely partially from rejecting inherited wrong previous values. 
Result of full state model also derived the similar conclusion.

Figure \ref{fig:indirectInherit} describes the result of indirect mentioned slot inherit analysis on the same
experiment. It indicates the MSP model tracked indirectly mentioned slots 612 times, 
occupying about 91\% of indirectly mentioned 
cases, while we cannot investigate this ability of other models for their black box properity. 
Although we did not found appropriate baselines to proof the superiority of MSP model in inheriting 
indirectly mentioned slots, its superiority could be verified indirectly. Of note, the MSP-full model 
achieved slightly better performance than the MSP-self model. The only difference between the two models is 
that the MSP-full model contains information of indirectly mentioned slots, while the MSP-self model does not. 
Therefore, we can reasonably infer that the performance improvement of the MSP-full model is from the incorporation of indirectly mentioned slots. It likely improves the model's 
ability to handle indirectly mentioned slots, resulting in the JGA improvement.

\subsubsection{Error Analysis}
\begin{table*}[t]
    \centering
    \begin{tabular}{llccccccccc}
    \toprule
    Slot & Slot Type & Accuracy & Precision & Recall & F1 & TP & TN & FP & FN & PLFP \\
    \midrule
    taxi-leaveat & span & 98.9\% & 81.6\%& 92.6\%& 86.8\%& 3.6\%& 95.4\%& 0.3\% & 0.5\%& 0.3\% \\
    taxi-destination& span& 98.1\%& 82.1\%& 92.5\%& 87.0\%& 6.4\%& 91.7\%& 0.5\%& 0.4\% & 1.0\%\\
    taxi-depature&span&97.9\%& 79.1\%& 93.0\%& 85.5\%& 6.1\%& 91.8\%& 0.5\%& 0.4\%& 1.2\%\\
    taxi-arriveby&span&99.1\%&84.4\%&87.1\%& 85.7\%& 2.7\%& 96.5\%&0.4\%& 0.3\%& 0.2\%\\
    restaurant-name&span&94.8\%&84.9\%&92.0\%&88.3\%&19.8\%&75.0\%&1.7\%&2.9\%&0.6\%\\
    hotel-name&span&95.8\%&87.9\%&90.7\%&89.3\%&17.7\%&78.1\%&1.8\%&1.9\%&0.5\%\\
    hotel-parking&categorical&95.4\%&82.4\%&87.4\%&84.8\%&12.9\%&82.5\%&1.9\%&2.5\%&0.2\%\\
    hotel-type&categorical&95.4\%&82.4\%&87.4\%&84.8\%&12.9\%&82.5\%&1.9\%&2.5\%&0.2\%\\
    attraction-name&span&93.1\%&72.0\%&75.2\%&73.6\%&9.6\%&83.6\%&3.2\%&2.9\%&0.8\%\\
    train-leaveat&span&97.1\%&82.9\%&97.0\%&89.4\%&12.4\%&84.6\%&0.4\%&1.5\%&1.0\%\\
    \bottomrule
    \end{tabular}
    \caption{Error distribution}
    \label{table:ErrorDistribution}
\end{table*}
At last, we investigated weaknesses of our model. Table \ref{table:ErrorDistribution} describes the 
slot-specific error distribution 
of a MultiWOZ 2.2 dataset experiment. We only showed ten slots whose F1 values are less than 90\% to save space. 
These slots are the main performance bottleneck of DST tasks. It is not surprising that most slots are 
span slots because finding an exact correct token span in a dialogue with hunderends of tokens 
is difficult. We found the model idenfity wrong slot values mainly because of FP predictions and FN predictions,
which is not surprising as well. However, the error distribution revealed that the performance of 
many slots lagged because of the high 
PLFP rate. More than half of the mistakes are PLFP in taxi-destination and taxi-departure slot, 
and over 30\% of mistakes are PLFP in the train-leaveat slot. Previous studies have noticed this 
phenomenon, but they did not analyze it \cite{tian-2021-AGDST}.

We detailly investigated the high PLFP rate problem in this study. It seems that most PLFP mistakes 
occur in cases that require the 
model to identify the correct value in several candidate values. For example, when a user says, 
``\textit{I need a train leaving after 19:45.}'' and the agent replies, 
``\textit{There is a train leaving at 21:00.}'', 
there are two candidate values for the train-leaveat slot, i.e., ``19:45'', and ``21:00''. We found 
our model may predict ``19:45'', rather than the correct ``21:00''. 
This result reflected that our model understands shallow semantic 
information because it extracted a time token span rather than a meaningless one. However, it still 
cannot understand the deep semantic information because its prediction was wrong.

\section{Conclusion}
This study proposed a model with a MSP to improve DST performance. The experimental results indicate our 
model reached new SOTA in DST tasks in MultiWOZ 2.1 and 2.2 datasets. Further experiments demonstrated
the MSP can be used as an addidtional component to improve the DST performance, and the 
MSP-based dialogue state update strategy is more effective than other common update strategies. 
Meanwhile, we quantitatively analyzed that our design indeed helps the model reject wrong values 
and track indirectly mentioned slots. However, our model still performs 
poorly in understanding deep semantic information. In the future study, we will integrate external 
grammar knowledge to improve the model's understanding ability in complex dialogue context.

\section*{Acknowledgements}
This research was supported by the National Key Research and Development Program of China (No. 2021ZD0204105),
Exploratory Research Project of Zhejiang Lab (No. 2022RC0AN01), and Open Research Projects of Zhejiang Lab 
(No. 2019KD0AD01/012).

\bibliographystyle{named}
\bibliography{MSP}

\begin{thebibliography}{}

\bibitem[\protect\citeauthoryear{Chao and Lane}{2019}]{chao-2019-BERTDST}
Guan-Lin Chao and Ian Lane.
\newblock {BERT-DST}: Scalable end-to-end dialogue state tracking with
  bidirectional encoder representations from transformer.
\newblock In {\em INTERSPEECH}, 2019.

\bibitem[\protect\citeauthoryear{Devlin \bgroup \em et al.\egroup
  }{2019}]{devlin-2019-BERT}
Jacob Devlin, Ming-Wei Chang, Kenton Lee, and Kristina Toutanova.
\newblock {BERT}: Pre-training of deep bidirectional transformers for language
  understanding.
\newblock In {\em Proceedings of the 2019 Conference of the North {A}merican
  Chapter of the Association for Computational Linguistics: Human Language
  Technologies, Volume 1 (Long and Short Papers)}, pages 4171--4186, June 2019.

\bibitem[\protect\citeauthoryear{Eric \bgroup \em et al.\egroup
  }{2020}]{eric-2020-multiwoz21}
Mihail Eric, Rahul Goel, Shachi Paul, Abhishek Sethi, Sanchit Agarwal, Shuyang
  Gao, Adarsh Kumar, Anuj Goyal, Peter Ku, and Dilek Hakkani-Tur.
\newblock {M}ulti{WOZ} 2.1: A consolidated multi-domain dialogue dataset with
  state corrections and state tracking baselines.
\newblock In {\em Proceedings of the 12th Language Resources and Evaluation
  Conference}, pages 422--428, May 2020.

\bibitem[\protect\citeauthoryear{Feng \bgroup \em et al.\egroup
  }{2021}]{feng-2021-sequence}
Yue Feng, Yang Wang, and Hang Li.
\newblock A sequence-to-sequence approach to dialogue state tracking.
\newblock In {\em Proceedings of the 59th Annual Meeting of the Association for
  Computational Linguistics and the 11th International Joint Conference on
  Natural Language Processing (Volume 1: Long Papers)}, pages 1714--1725,
  August 2021.

\bibitem[\protect\citeauthoryear{Gao \bgroup \em et al.\egroup
  }{2019}]{gao-2019-DSTReadComprehension}
Shuyang Gao, Abhishek Sethi, Sanchit Agarwal, Tagyoung Chung, and Dilek
  Hakkani-Tur.
\newblock Dialog state tracking: A neural reading comprehension approach.
\newblock In {\em Proceedings of the 20th Annual SIGdial Meeting on Discourse
  and Dialogue}, pages 264--273, September 2019.

\bibitem[\protect\citeauthoryear{Heck \bgroup \em et al.\egroup
  }{2020}]{heck-2020-trippy}
Michael Heck, Carel van Niekerk, Nurul Lubis, Christian Geishauser, Hsien-Chin
  Lin, Marco Moresi, and Milica Gasic.
\newblock Trippy: A triple copy strategy for value independent neural dialog
  state tracking.
\newblock In {\em Proceedings of the 21th Annual Meeting of the Special
  Interest Group on Discourse and Dialogue}, pages 35--44, July 2020.

\bibitem[\protect\citeauthoryear{Hosseini-Asl \bgroup \em et al.\egroup
  }{2020}]{hosseini-Asl-2020-SimpleTOD}
Ehsan Hosseini-Asl, Bryan McCann, Chien-Sheng Wu, Semih Yavuz, and Richard
  Socher.
\newblock A simple language model for task-oriented dialogue.
\newblock In H.~Larochelle, M.~Ranzato, R.~Hadsell, M.~F. Balcan, and H.~Lin,
  editors, {\em Advances in Neural Information Processing Systems}, volume~33,
  pages 20179--20191, 2020.

\bibitem[\protect\citeauthoryear{Kim \bgroup \em et al.\egroup
  }{2018}]{kim-2018-bilinear}
Jin-Hwa Kim, Jaehyun Jun, and Byoung-Tak Zhang.
\newblock Bilinear attention networks.
\newblock In {\em Proceedings of the 32nd International Conference on Neural
  Information Processing Systems}, NIPS'18, page 1571–1581, 2018.

\bibitem[\protect\citeauthoryear{Kim \bgroup \em et al.\egroup
  }{2020}]{kim-2020-OverwriteMemory}
Sungdong Kim, Sohee Yang, Gyuwan Kim, and Sang-Woo Lee.
\newblock Efficient dialogue state tracking by selectively overwriting memory.
\newblock In {\em Proceedings of the 58th Annual Meeting of the Association for
  Computational Linguistics}, pages 567--582, July 2020.

\bibitem[\protect\citeauthoryear{Kingma and Ba}{2015}]{kingma-2014-adam}
Diederik~P Kingma and Jimmy Ba.
\newblock Adam: A method for stochastic optimization.
\newblock {\em International Conference on Learning Representations}, 2015.

\bibitem[\protect\citeauthoryear{Lee \bgroup \em et al.\egroup
  }{2019}]{lee-2019-sumbt}
Hwaran Lee, Jinsik Lee, and Tae-Yoon Kim.
\newblock {SUMBT}: Slot-utterance matching for universal and scalable belief
  tracking.
\newblock In {\em Proceedings of the 57th Annual Meeting of the Association for
  Computational Linguistics}, pages 5478--5483, July 2019.

\bibitem[\protect\citeauthoryear{Li \bgroup \em et al.\egroup
  }{2021}]{li-2020-coco}
Shiyang Li, Semih Yavuz, Kazuma Hashimoto, Jia Li, Tong Niu, Nazneen Rajani,
  Xifeng Yan, Yingbo Zhou, and Caiming Xiong.
\newblock Coco: Controllable counterfactuals for evaluating dialogue state
  trackers.
\newblock {\em International Conference on Learning Representations}, 2021.

\bibitem[\protect\citeauthoryear{Liu \bgroup \em et al.\egroup
  }{2021}]{liu-2021-promptSurvey}
Pengfei Liu, Weizhe Yuan, Jinlan Fu, Zhengbao Jiang, Hiroaki Hayashi, and
  Graham Neubig.
\newblock Pre-train, prompt, and predict: A systematic survey of prompting
  methods in natural language processing.
\newblock {\em arXiv preprint arXiv:2107.13586}, 2021.

\bibitem[\protect\citeauthoryear{Manotumruksa \bgroup \em et al.\egroup
  }{2021}]{manotumruksa-2021-turnBaseLoss}
Jarana Manotumruksa, Jeff Dalton, Edgar Meij, and Emine Yilmaz.
\newblock Improving dialogue state tracking with turn-based loss function and
  sequential data augmentation.
\newblock In {\em Findings of the Association for Computational Linguistics:
  EMNLP 2021}, pages 1674--1683, November 2021.

\bibitem[\protect\citeauthoryear{Mehri \bgroup \em et al.\egroup
  }{2020}]{mehri-2020-DialoGLUE}
S.~Mehri, M.~Eric, and D.~Hakkani-Tur.
\newblock Dialoglue: A natural language understanding benchmark for
  task-oriented dialogue.
\newblock {\em arXiv preprint arXiv:2009.13570}, 2020.

\bibitem[\protect\citeauthoryear{Ni \bgroup \em et al.\egroup
  }{2021}]{ni-2021-DSTReview}
Jinjie Ni, Tom Young, Vlad Pandelea, Fuzhao Xue, Vinay Adiga, and Erik Cambria.
\newblock Recent advances in deep learning-based dialogue systems.
\newblock {\em arXiv preprint arXiv:2105.04387}, 2021.

\bibitem[\protect\citeauthoryear{Nouri and
  Hosseini-Asl}{2018}]{nouri-2018-ScalableDST}
Elnaz Nouri and Ehsan Hosseini-Asl.
\newblock Toward scalable neural dialogue state tracking.
\newblock In {\em NeurIPS 2018, 2nd Conversational AI workshop}, 2018.

\bibitem[\protect\citeauthoryear{Song \bgroup \em et al.\egroup
  }{2021}]{song-2021-DSTDataAugment}
Xiaohui Song, Liangjun Zang, and Songlin Hu.
\newblock Data augmentation for copy-mechanism in dialogue state tracking.
\newblock In {\em International Conference on Computational Science}, pages
  736--749. Springer, 2021.

\bibitem[\protect\citeauthoryear{Summerville \bgroup \em et al.\egroup
  }{2020}]{summerville-2020-DSTDataAugment}
Adam Summerville, Jordan Hashemi, James Ryan, and William Ferguson.
\newblock How to tame your data: Data augmentation for dialog state tracking.
\newblock In {\em Proceedings of the 2nd Workshop on Natural Language
  Processing for Conversational AI}, pages 32--37, July 2020.

\bibitem[\protect\citeauthoryear{Tian \bgroup \em et al.\egroup
  }{2021}]{tian-2021-AGDST}
Xin Tian, Liankai Huang, Yingzhan Lin, Siqi Bao, Huang He, Yunyi Yang, Hua Wu,
  Fan Wang, and Shuqi Sun.
\newblock Amendable generation for dialogue state tracking.
\newblock In {\em Proceedings of the 3rd Workshop on Natural Language
  Processing for Conversational AI}, pages 80--92, November 2021.

\bibitem[\protect\citeauthoryear{Townsend \bgroup \em et al.\egroup
  }{2001}]{townsend-2001-sentence}
David~J Townsend, Thomas~G Bever, Thomas~G Bever, et~al.
\newblock {\em Sentence comprehension: The integration of habits and rules}.
\newblock MIT Press, 2001.

\bibitem[\protect\citeauthoryear{Wen \bgroup \em et al.\egroup
  }{2017}]{wen-2017-WOZ2}
Tsung-Hsien Wen, David Vandyke, Nikola Mrk{\v{s}}i{\'c}, Milica
  Ga{\v{s}}i{\'c}, Lina~M. Rojas-Barahona, Pei-Hao Su, Stefan Ultes, and Steve
  Young.
\newblock A network-based end-to-end trainable task-oriented dialogue system.
\newblock In {\em Proceedings of the 15th Conference of the {E}uropean Chapter
  of the Association for Computational Linguistics: Volume 1, Long Papers},
  pages 438--449, April 2017.

\bibitem[\protect\citeauthoryear{Wu \bgroup \em et al.\egroup
  }{2016}]{wu-2016-google}
Yonghui Wu, Mike Schuster, Zhifeng Chen, Quoc~V Le, Mohammad Norouzi, Wolfgang
  Macherey, Maxim Krikun, Yuan Cao, Qin Gao, Klaus Macherey, et~al.
\newblock Google's neural machine translation system: Bridging the gap between
  human and machine translation.
\newblock {\em arXiv preprint arXiv:1609.08144}, 2016.

\bibitem[\protect\citeauthoryear{Wu \bgroup \em et al.\egroup
  }{2019}]{wu-2019-TRADE}
Chien-Sheng Wu, Andrea Madotto, Ehsan Hosseini-Asl, Caiming Xiong, Richard
  Socher, and Pascale Fung.
\newblock Transferable multi-domain state generator for task-oriented dialogue
  systems.
\newblock In {\em Proceedings of the 57th Annual Meeting of the Association for
  Computational Linguistics (Volume 1: Long Papers)}, 2019.

\bibitem[\protect\citeauthoryear{Zang \bgroup \em et al.\egroup
  }{2020}]{zang-2020-MultiWOZ22}
Xiaoxue Zang, Abhinav Rastogi, Srinivas Sunkara, Raghav Gupta, Jianguo Zhang,
  and Jindong Chen.
\newblock {M}ulti{WOZ} 2.2 : A dialogue dataset with additional annotation
  corrections and state tracking baselines.
\newblock In {\em Proceedings of the 2nd Workshop on Natural Language
  Processing for Conversational AI}, pages 109--117, July 2020.

\bibitem[\protect\citeauthoryear{Zhang \bgroup \em et al.\egroup
  }{2020}]{zhang-2020-dual}
Jianguo Zhang, Kazuma Hashimoto, Chien-Sheng Wu, Yao Wang, Philip Yu, Richard
  Socher, and Caiming Xiong.
\newblock Find or classify? dual strategy for slot-value predictions on
  multi-domain dialog state tracking.
\newblock In {\em Proceedings of the Ninth Joint Conference on Lexical and
  Computational Semantics}, pages 154--167, December 2020.

\bibitem[\protect\citeauthoryear{Zhao \bgroup \em et al.\egroup
  }{2021}]{zhao-2021-effectiveSequence}
Jeffrey Zhao, Mahdis Mahdieh, Ye~Zhang, Yuan Cao, and Yonghui Wu.
\newblock Effective sequence-to-sequence dialogue state tracking.
\newblock In {\em Proceedings of the 2021 Conference on Empirical Methods in
  Natural Language Processing}, pages 7486--7493, November 2021.

\bibitem[\protect\citeauthoryear{Zhou and Small}{2019}]{zhou-2019-DSTQA}
Li~Zhou and Kevin Small.
\newblock Multi-domain dialogue state tracking as dynamic knowledge graph
  enhanced question answering.
\newblock In {\em NeurIPS Workshop on Conversational AI}, 2019.

\end{thebibliography}

\end{document}